\documentclass[letterpaper, 10 pt, conference]{ieeeconf}  

\IEEEoverridecommandlockouts                              

\usepackage{amsfonts, dsfont, mathtools, amsmath, amssymb, bbold, mathrsfs, bm, upgreek}
\usepackage{graphicx, float, epsfig, color, psfrag, adjustbox, enumerate}
\usepackage{subcaption}
\usepackage{nicefrac}
\usepackage{cleveref, refcount, url}
\usepackage[noadjust]{cite}
\usepackage[usenames,dvipsnames,svgnames,table]{xcolor}
\usepackage{algorithm}
\usepackage{algpseudocode}

\algrenewcommand\alglinenumber[1]{#1:}
\usepackage{multirow}
\usepackage{multicol}
\usepackage{tikz}
\usetikzlibrary{positioning,arrows}
\usepackage{arydshln}

\title{\LARGE \bf
Controlling a CyberOctopus Soft Arm with Muscle-like Actuation
}

\author{Heng-Sheng Chang$^{1,2}$, Udit Halder$^2$, Ekaterina Gribkova$^{3}$, Arman Tekinalp$^{1}$,\\
 [5pt] Noel Naughton$^{1}$, Mattia Gazzola$^{1,4,5}$, Prashant G. Mehta$^{1,2,3}$
\thanks{We gratefully acknowledge financial support from ONR MURI N00014-19-1-2373, NSF/USDA $\#$2019-67021-28989, and NSF EFRI C3 SoRo $\#$1830881. We also acknowledge computing resources provided  by the Blue Waters project (OCI- 0725070, ACI-1238993), a joint effort of the University of Illinois at Urbana-Champaign and its National Center for Supercomputing Applications, and the Extreme Science and Engineering Discovery Environment (XSEDE) Stampede2 (ACI-1548562) at the Texas Advanced Computing Center (TACC) through allocation TG-MCB190004.}%
\thanks{$^{1}$Department of Mechanical Science and Engineering, $^{2}$Coordinated Science Laboratory, $^3$Neuroscience Program, $^{4}$National Center for Supercomputing Applications, \&  $^{5}$Carl R. Woese Institute for Genomic Biology, University of Illinois at Urbana-Champaign. Corresponding e-mail: {\tt\small udit@illinois.edu}}
}

\def\R{{\mathds{R}}}

\def\0{{\mathbb{0}}}
\def\1{{\mathds{1}}}

\def\a{{\mathbf{a}}}
\def\b{{\mathbf{b}}}

\newcommand{\norm}[1]{\left\lVert#1\right\rVert}

\newcommand{\abs}[1]{\left| #1 \right|}

\definecolor{db}{RGB}{23,20,119}
\definecolor{dg}{RGB}{2,101,15}

\newtheorem{corollary}{Corollary}[section]

\newtheorem{proposition}{Proposition}[section]
\newtheorem{definition}{Definition}
\newtheorem{remark}{Remark}

\usepackage{upgreek}
\newcommand{\dif}{\mathrm{d}}
\newcommand{\transpose}{\intercal}
\newcommand{\set}[1]{\left\{#1\right\}}

\newcommand{\material}[1]{
	\ifthenelse{\equal{#1}{\kappa}}{\upkappa}{
	\ifthenelse{\equal{#1}{\nu}}{\upnu}{
	\ifthenelse{\equal{#1}{\omega}}{\upomega}{
	\ifthenelse{\equal{#1}{\sigma}}{\upsigma}{
	\ifthenelse{\equal{#1}{\theta}}{\uptheta}{
	\mathsf{#1}}}}}}
}

\newcommand{\intrinsic}{\circ}

\newcommand{\deformations}{w}
\newcommand{\states}{q}
\newcommand{\momentums}{p}
\newcommand{\costates}{\lambda}
\newcommand{\Hamiltonian}{\mathcal{H}}
\newcommand{\potential}{\mathcal{V}}
\newcommand{\kinetic}{\mathcal{T}}
\newcommand{\mass}{M}
\newcommand{\controlHamiltonian}{H}

\renewcommand{\a}{\mathsf{a}}
\renewcommand{\b}{\mathsf{b}}

\newcommand{\muscle}{\text{m}}
\newcommand{\longitudinalmuscle}{\text{LM}}
\newcommand{\transversemuscle}{\text{TM}}
\newcommand{\Gmuscle}{\mathsf{G}}

\newcommand{\LMt}{\text{LM}_\text{t}}
\newcommand{\LMb}{\text{LM}_\text{b}}

\graphicspath{{figures/}}

\begin{document}
\bstctlcite{BSTcontrol} 
\maketitle
\thispagestyle{empty}
\pagestyle{empty}


\begin{abstract}
This paper presents an application of the energy shaping methodology to control a flexible, elastic Cosserat rod model of a single octopus arm.  The novel contributions of this work are two-fold: (i) a control-oriented modeling of the anatomically realistic internal muscular architecture of an octopus arm; and (ii) the integration of these muscle models into the energy shaping control methodology. The control-oriented modeling takes inspiration in equal parts from theories of nonlinear elasticity and energy shaping control.  By introducing a stored energy function for muscles, the difficulties associated with explicitly solving the matching conditions of the energy shaping methodology are avoided.  The overall control design problem is posed as a bilevel optimization problem.  Its solution is obtained through iterative algorithms. The methodology is numerically implemented and demonstrated in a full-scale dynamic simulation environment {\em Elastica}. Two bio-inspired numerical experiments involving the control of octopus arms are reported.
\end{abstract}


\section{Introduction} \label{sec:intro}
Research interest in soft robotic manipulators comes from the potential capability of soft manipulators to perform complex tasks in an unstructured environment and safely around humans~\cite{rus2015design, singh2017constrained, della2018dynamic}.  Bio-inspiration is often provided by soft-bodied creatures, such as octopuses, that have evolved to solve complex motion control problems like reaching, grasping, fetching, crawling, or swimming. The exceptional coordination abilities of these marine animals have naturally motivated efforts to gain a deeper understanding of the biophysical principles underlying their distributed neuromuscular control.


This paper is a continuation of our prior work \cite{chang2020energy} where an energy shaping methodology was introduced for a fully actuated Cosserat rod model of an octopus arm. The methodology was applied to solve motion problems, e.g. reaching and grasping, inspired by experiments involving octopus arms~\cite{levy2017motor}. A major limitation of this prior work was that the muscle actuator constraints were ignored. 
In this paper, we have worked closely with the biologists on the team to incorporate anatomically correct muscular architecture of longitudinal, transverse, and oblique muscles in octopus arms~\cite{kier2007arrangement, kier2016musculature, feinstein2011functional, nesher2019synaptic}.


From a control theoretic perspective of energy shaping, the muscle constraints limit the energy landscape that can possibly be `shaped' by the use of control.  In the energy shaping literature, a mathematical characterization of achievable energy is given in terms of the so-called \textit{matching conditions}~\cite{ortega2002stabilization, blankenstein2002matching}. However, solving these conditions is a formidable task~\cite{bloch2000controlled, auckly2000control}.  Our preliminary attempts at directly applying the energy shaping methodology ran into this challenging problem.   

The key idea of this paper is to isolate and model the conservative parts of the muscle forces using the formalism of {\em stored energy function} borrowed from nonlinear elasticity~\cite{antman1995nonlinear, healey2006straightforward}.  This has an advantage in the sense that it circumvents the need to explicitly solve the matching conditions.  The specific contributions are as follows:

\smallskip

\begin{figure*}[!t]
	\centering
	\includegraphics[width=\textwidth, trim = {30 230 30 70}, clip = true]{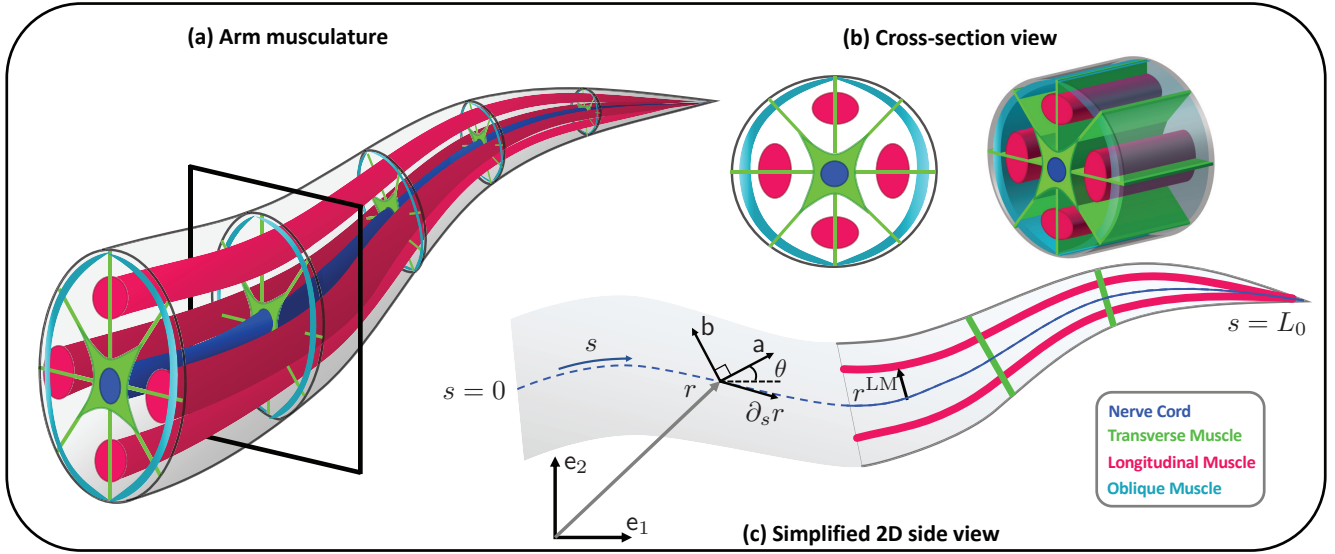}
	\caption{An octopus arm is modeled as a Cosserat rod -- (a) a simplified 3D view of the internal musculature, (b) a cross-section of the arm showing organization of different kinds of muscles, and (c) a 2D model of the arm where we only consider effects of longitudinal and transverse muscles.}
	\label{fig:model}
	\vspace{-15pt}
\end{figure*}

\noindent \textbf{1. Control-oriented modeling of muscles:} Muscles provide internal (tension) forces.  The modeling of a muscle is based upon the Hill's model~\cite{hill1938heat, winter2009biomechanics, audu1985influence, yekutieli2005dynamic1} that prescribes a force-length (conservative) and a force-velocity (dissipative) relationship for an activated muscle.  Our contribution is to model the conservative muscle force in terms of the stored energy function of nonlinear elasticity.  For the longitudinal and transverse muscles of an octopus arm, explicit expressions of the stored energy function are derived based upon first principle modeling and published force-length characteristics. 

The inspiration for such a model comes in equal part from the hyperelastic rod theory and from the energy shaping control theory. Similar to the elasticity in Cosserat rod theory~\cite{antman1995nonlinear}, the control terms on account of muscles are modeled as Hamiltonian vector fields. 

\smallskip

\noindent \textbf{2. Application of energy shaping for muscular arm control:} The control problem is to activate the muscles to solve motion control tasks, e.g. reaching and grasping. The control problem is posed and solved as a bilevel optimization problem. The upper level problem seeks to find the optimal muscle activation subject to the equilibrium constraints that are shown to arise as a result of solving a lower level optimization problem.  For the bilevel optimization problem, necessary conditions of optimality are derived by an application of the maximum principle of the optimal control theory. This also leads to iterative algorithms to compute the control.  Conditions are given for the asymptotic stability of the rod equilibrium obtained using the control.

\smallskip

The dynamics of the arm together with the muscle actuation model are simulated in the computational platform \textit{Elastica} \cite{gazzola2018forward, zhang2019modeling, naughton2021elastica}. The control methodology is implemented and demonstrated for reaching and grasping tasks.

The remainder of this paper is organized as follows.  An anatomical overview of the arm musculature, the planar Cosserat rod modeling, and the control objectives appear in Sec.~\ref{sec:model}. The two main contributions of this paper appear in Sec.~\ref{sec:muscle_model} on control-oriented muscle modeling, and in Sec.~\ref{sec:control} on application of energy shaping. Numerical experiments are reported in Sec.~\ref{sec:numerics} and conclusions are given in Sec.~\ref{sec:conclusion}.

\section{Modeling}
\label{sec:model}

\subsection{Physiology of a muscular octopus arm}

An octopus arm is composed of a central axial nerve cord which is surrounded by densely packed muscle and connective tissues.  Organizationally, the arm muscles are of three types~\cite{kier2007arrangement, feinstein2011functional, kier2016musculature}:
\begin{enumerate}
\item Longitudinal muscles run parallel to the axial nerve cord.  These muscles are responsible for bending and shortening of the arm. 
\item Transverse muscles surround the nerve cord.  These muscles are used to lengthen the arm and provide support in active bending.
\item Oblique muscles are the outermost helical muscle fibers. These muscles provide twist.  
\end{enumerate} 
A simplified 3D model depicting the muscular organization appears in Fig.~\ref{fig:model}a. In this paper, we will focus on the planar case.

\subsection{Modeling a single arm as a Cosserat rod} 

\noindent 
\textbf{Kinematics:} Consider a Cosserat rod in a plane spanned by the fixed orthonormal basis $\set{\mathsf{e}_1,\mathsf{e}_2}$. The rod is a one-dimensional continuum deformable object with rest length $L_0$.  There are two independent coordinates: time $t\in\R$ and the arc-length coordinate $s \in [0, L_0]$ of the center line.  The \textit{configuration} of the rod is described by the function 
\[
\states (s, t) := (r(s,t), \theta (s, t))
\]  
where $r=(x,y)\in\R^2$ denotes the position vector of the center line, and the angle $\theta\in\R$ defines a material frame spanned by the orthonormal basis $\{\a,\b\}$ where (see Fig.~\ref{fig:model}c)
\[
\a = \cos \theta \,\mathsf{e}_1 + \sin \theta \, \mathsf{e}_2, ~ \b = -\sin \theta \, \mathsf{e}_1 + \cos \theta \, \mathsf{e}_2
\] 
The \textit{deformations} or \textit{strains} -- stretch ($\nu_1$), shear ($\nu_2$), and curvature ($\kappa$) -- denoted together as $\deformations = (\nu_1, \nu_2, \kappa)$ and  are defined through the kinematic relationship 
\begin{equation}
		\partial_s\states = \begin{bmatrix}		\partial_s r \\ \partial_s \theta 	\end{bmatrix} = 
			\begin{bmatrix}
				\nu_1 \a +\nu_2 \b\\
				\kappa
			\end{bmatrix} =: g(q, w)
			\label{eq:state_kinematics}
\end{equation}
where $\partial_s := \frac{\partial}{\partial s}$ is the partial derivative with respect to $s$. A rod is said to be inextensible if $\nu_1\equiv1$ and un-shearable if $\nu_2 \equiv 0$.   

\medskip

\noindent 
\textbf{Dynamics:} The Cosserat rod is expressed as a Hamiltonian control system. For this purpose, let $\mass = \text{diag}(\rho A, \rho A, \rho I)$ be the inertia matrix where $\rho$ is the material density, $A$ and $I$ are the cross sectional area and the second moment of area, respectively. 
The state of the Hamiltonian system is $(q(s,t),p(s,t))$ where \textit{momentum} $\momentums := \mass \partial_t\states$ and $\partial_t :=\frac{\partial }{\partial t}$ is the partial derivative with respect to $t$.  To obtain the equations of motions, one needs to specify the kinetic energy and the potential energy.  The kinetic energy of the rod is
\begin{align*}
\kinetic= \frac{1}{2} \int_0^{L_0} \left(\rho A ( (\partial_t x)^2 + (\partial_t y)^2) +  \rho I (\partial_t \theta)^2 \right) \dif s
\end{align*}
The potential energy of a {\em hyperelastic} Cosserat rod model is given by\footnote{Here the superscript $e$ stands for elastic potential energy} 
\begin{align}
\potential^e &= \int_0^{L_0} W^e\left(\deformations \right) \dif s 
\label{eq:potential}
\end{align}
where $W^e(w) = W^e(\nu_1,\nu_2,\kappa)$ is referred to as the elastic {\em stored energy function}. The simplest model of the stored energy function is of the quadratic form
	\begin{equation}
		W^e = \frac{1}{2} \left( EA(\nu_1-\nu_1^\intrinsic)^2 + GA(\nu_2-\nu_2^\intrinsic)^2+ EI(\kappa-\kappa^\intrinsic)^2 \right) \nonumber
	\end{equation}
where $E$ and $G$ are the material Young's and shear moduli, 
and $\nu_1^\intrinsic$, $\nu_2^\intrinsic$, $\kappa^\intrinsic$ are the intrinsic strains that determine the rod's shape at rest. If $\nu_1^\intrinsic \equiv 1$, $\nu_2^\intrinsic \equiv 0$, $\kappa^\intrinsic \equiv 0$, then the rest configuration is a straight rod of length $L_0$.  

\medskip

Thus, the overall dynamic model is expressed as a Hamiltonian control system with damping:
 \begin{align}
\begin{split}
\frac{\dif \states}{\dif t} &= \frac{\delta \kinetic}{\delta \momentums} \\
\frac{\dif \momentums}{\dif t} &= - \frac{\delta \potential^e}{\delta \states} - \gamma M^{-1} \momentums
+ \sum\limits_{\muscle} \Gmuscle^\muscle(\states) u^{\muscle} 
\end{split}
\label{eq:control_system}
\end{align}
where $u^\muscle=u^\muscle (s, t) \in [0, 1]$ is the muscle activation\footnote{Physically, the control $u^\muscle(s,t)$ is related to the neuronal stimulation, such as the firing frequency of the motor neuron innervating the muscle fiber at $s$ at time $t$. The bound on $u^\muscle$ is a manifestation of maximum firing frequency of a motor neuron.} and serves as the control input for our purposes. The operators $\Gmuscle^\muscle$ model how a (internal) muscle actuation $u^\muscle$ translates to resultant (external) forces and couples that move the Cosserat rod. The modeling of $\Gmuscle^\muscle$ is the subject of the next section.  The linear damping term models the inherent viscoelasticity in the arm \cite{gazzola2018forward} where $\gamma > 0$ is a damping coefficient. Model~\eqref{eq:control_system} is accompanied by suitable initial and boundary conditions. 

\medskip

\begin{remark}
For a given $W^e$, the internal elastic forces and couples are given by the \textit{stress-strain relationship}
\begin{align*}
n_i = \frac{\partial W^e}{\partial \nu_i}, i = 1, 2, ~~ m = \frac{\partial W^e}{\partial \kappa}
\end{align*}
where $n_1, n_2$ are the components of the internal force $(n)$ in the material frame, i.e. $n = n_1 \mathsf{a} + n_2 \mathsf{b}$, and $m$ is the internal couple.  
Specifically, the conservative forces on the righthand-side of~\eqref{eq:control_system} are as follows \cite{antman1995nonlinear}:
\begin{equation*}
- \frac{\delta \potential^e}{\delta \states} =\begin{bmatrix}
\partial_s\left( \begin{pmatrix*}[r] \cos \theta & - \sin \theta \\ \sin \theta & \cos \theta \end{pmatrix*}
\begin{pmatrix*} n_1  \\ n_2  \end{pmatrix*} \right) \\
\partial_s m + \nu_1n_2  -\nu_2n_1
\end{bmatrix}
\end{equation*}
The quadratic model of the stored energy means that the stress-strain relationship is linear.  For this reason, the quadratic model of the stored energy function is referred to as linear elasticity.  However, elastic characteristics of soft tissue can be nonlinear~\cite{tramacere2014structure}.    
\end{remark}


\medskip

\noindent \textbf{Control objective:} Octopus arms are capable of a variety of manipulation tasks \cite{levy2017motor}. Inspired by this, we set some control problems to be solved by our model arm. In this work, the control objective is to design stabilizing distributed muscle activations so that -- (i) the tip reaches a target point, and (ii) the arm wraps around an object to grasp it.

\section{Control-oriented modeling of muscles} \label{sec:muscle_model}

In this section, the control-oriented modeling of internal muscles is presented. The main results of this section are given in Proposition~\ref{prop:muscle_energy} and Corollary~\ref{cor:closed_loop_hamiltonian}.  A reader more interested in the control problem may skip ahead to Sec.~\ref{sec:muscle energy} which is followed by Sec.~\ref{sec:control} on control methodology. 


\subsection{Muscle geometry}\label{sec:muscle_geometry}

Since we consider planar movement of the arm, the following muscles are most relevant (see Fig.~\ref{fig:model}c)
\begin{enumerate}
\item The top longitudinal muscle denoted as $\LMt$,
\item The bottom longitudinal muscle denoted as $\LMb$, 
\item The transverse muscles in the middle denoted as $\transversemuscle$.
\end{enumerate}
For a generic muscle $\muscle \in \{ \LMt, \LMb, \transversemuscle\}$, the vector   
\[
r^\muscle=r^\muscle_1\mathsf{a}+r^\muscle_2\mathsf{b}
\] 
specifies its position with respect to the centerline, as depicted in Fig.~\ref{fig:model}c.  For these three muscle groups, coordinates $(r^\muscle_1,r^\muscle_2)$ are reported in Table~\ref{tab:model}.

\subsection{Forces from muscle actuation}\label{sec:muscle_force_model}
When activated, a muscle provides an internal distributed contraction force
\begin{equation*}
	n^\muscle =n^\muscle_1 \mathsf{a}+n^\muscle_2  \mathsf{b}
\end{equation*}
and because of their (off centerline) geometric arrangement, the two longitudinal muscles also provide a couple 
\begin{equation*}
	m^\muscle=(r^\muscle \times n^\muscle)\cdot(\mathsf{e}_1 \times \mathsf{e}_2) =  r_1^\muscle n_2^\muscle-r_2^\muscle n_1^\muscle
\end{equation*}
where `$\cdot$' and `$\times$' denote the usual vector dot and cross product, respectively. This yields the following model for a single muscle:
\begin{equation}
	\Gmuscle^\muscle (\states) u^\muscle =\begin{bmatrix}
		\partial_s\left( 
			\begin{pmatrix*}[r] \cos \theta & - \sin \theta \\ \sin \theta & \cos \theta \end{pmatrix*}
			\begin{pmatrix*} n_1^\muscle  \\ n_2^\muscle  \end{pmatrix*} 
		\right) \\
		\partial_s m^\muscle  + \nu_1n_2^\muscle  -\nu_2n_1^\muscle 
	\end{bmatrix}
\label{eq:muscle_model}
\end{equation}
It remains to prescribe a model for the contraction force $n^\muscle$ as a function of the activation $u^\muscle$.  This is the subject of the following subsections. 

\subsection{Hill's lumped model of a muscle} \label{sec:hills_3element}

\begin{figure}[t]
	\centering
	\includegraphics[width = \columnwidth, trim = {5 120 95 120}, clip = true]{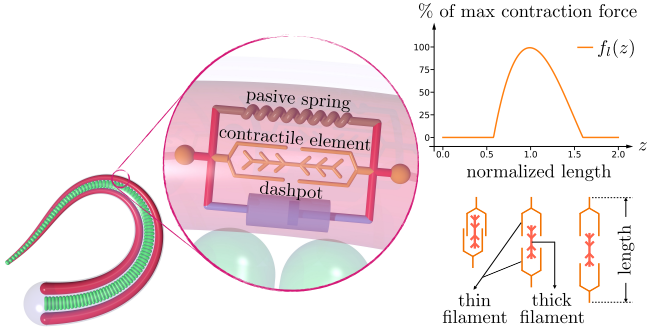}
	\caption{Hill's muscle model
	: The three parallel elements are shown in the inset. The force-length relationship of the active contractile element is shown on the right.
	}
	\label{fig:hills_model}
	\vspace{-10pt}
\end{figure}

In the Hill's muscle model~\cite{hill1938heat, audu1985influence, fung1996biomechanics, winter2009biomechanics}, the tension produced by muscular activation is modeled using three parallel lumped elements (Fig.~\ref{fig:hills_model}): an active contractile element, a passive spring element, and a damping element.  When a muscle is activated (i.e., $u^\muscle>0$), the active contractile element produces a tension force which scales linearly with activation $u^\muscle\in[0,1]$, and is described through a force-length relationship $f_l(\cdot)$, such as the one  depicted in Fig.~\ref{fig:hills_model}.  Maximum forces are produced when the muscle is in its rest length configuration, and decreases as the muscle contracts or elongates.
The bio-physics of this force-length relationship is explained by the sliding filament theory of individual muscle fibres~\cite{fung1996biomechanics, winter2009biomechanics}.  The passive spring element accounts for the elasticity of the muscle. There are also dissipative effects which are accounted for through so-called force-velocity relationship of the muscles. The following notation is adopted to model the force-length relationship.

\smallskip

\begin{definition}
Let $f_l : \R^+ \rightarrow \R^+$ denote the force-length relationship. Define also $F_l : \R^+ \rightarrow \R^+$, $F_l(z) := \int_0^z f_l(z) ~ \dif z$, i.e. $F_l '(z) = f_l(z)$, where the prime indicates the derivative with respect to the argument.
\end{definition}

\smallskip
\begin{remark}
We note here that a particular form of the function $f_l (\cdot)$ is not instrumental to the muscle modeling and control design, and can be replaced by an experimentally obtained relationship.  The specific form of the force-length relationship used here is a 3rd degree polynomial fit to experimental data for squid tentacle muscles~\cite[Fig. 6]{kier2002fast}. The formula appears in Sec.~\ref{sec:numerics}.
\end{remark}
\smallskip
\begin{remark}
Similar to \cite{ekeberg1993combined, yekutieli2005dynamic1}, in this paper the other two elements -- passive spring and the dashpot -- are not explicitly modeled. The effects of these elements are considered to be assimilated in the inherent viscoelasticity of the Cosserat rod, i.e. the gradient of $\potential^e$ term and the damping term in \eqref{eq:control_system}. 
\end{remark}

\subsection{Muscle model adaptation to Cosserat rod formalism} \label{sec:hills_continuum}

Here we adapt the Hill's active force model as follows. For a generic muscle $\muscle$, the magnitude of the active internal force is
\begin{align}
\abs{n^{\muscle}} &= n^\muscle_{\text{max}} A^\muscle(s) f_l(\nu^\muscle(s)) u^\muscle(s,t)
\label{eq:muscle_force}
\end{align}
where $A^\muscle$ is 
the arm cross sectional area occupied by the muscle, $n^\muscle_{\text{max}}$ is the maximum producible force per unit area.  In the context of this paper, the appropriate `length' is the stretch strain $\nu^\muscle$ of the muscle.  It is a function of the deformation $w$.  
In the following, we describe the models -- expressions for $n^\muscle, m^\muscle, \nu^\muscle$ for the two types of muscles:   

\medskip

\noindent \textbf{Longitudinal muscles:} The top and the bottom longitudinal muscles (LM) are positioned at a distance $\phi^\longitudinalmuscle (s)$ away from the rod center line.  Hence, the position vector of a longitudinal muscle is given as $r + r^\longitudinalmuscle$, where $r^\longitudinalmuscle=\pm \phi^\longitudinalmuscle\mathsf{b}$.  (The sign is positive for the top LM and negative for the bottom LM.)  Taking the spatial derivative of the muscle position $r + r^\longitudinalmuscle$, the local stretch strain is obtained as
\[
\nu^\longitudinalmuscle = \nu_1 \mp \phi^\longitudinalmuscle \kappa
\]
When a longitudinal muscle is activated, it generates the contraction force $n^\longitudinalmuscle$ along the longitudinal direction $\mathsf{a}$ and also produces the couple $m^\longitudinalmuscle$.  The expressions for these are obtained as
\[
n^\longitudinalmuscle =\abs{n^\longitudinalmuscle} \mathsf{a},\quad  m^\longitudinalmuscle =\mp \phi^\longitudinalmuscle \abs{n^\longitudinalmuscle}
\]
where $\abs{n^\longitudinalmuscle}$ is defined using~\eqref{eq:muscle_force} with $\muscle=\longitudinalmuscle$ and $\nu^\muscle$ given by $\nu^\longitudinalmuscle$.  
The couple $m^\longitudinalmuscle$ bends the arm locally which is believed to be one of the most important functionalities of the longitudinal muscle. By symmetry, the parameters $A^\muscle$ and $n^\muscle_{\text{max}}$ are considered to be the same for both the $\LMt$ and $\LMb$. 

\medskip

\noindent \textbf{Transverse muscles:}
When a transverse muscle (TM) contracts, it causes the cross sectional area to shrink.  Owing to the constancy of volume (tissue incompressibility), the arm extends.  Such a behavior is modeled as follows:
\[
n^\transversemuscle=- \abs{n^\transversemuscle} \mathsf{a},\quad m^\transversemuscle = 0
\]
where the couple is zero because the transverse muscles  surround the axial nerve cord, i.e. $r^\transversemuscle = 0$. To capture the physics of inverse relationship between the stretch strain of the arm and that of the transverse muscles, we model the local stretch strain of transverse muscle as 
\[
\nu^\transversemuscle = \tfrac{\nu_1^\circ}{\nu_1} \approx 2  - \tfrac{\nu_1}{\nu_1^\circ}
\] 
For simplicity, we adopt a linear approximation at $\nu_1 = \nu_1^\circ$.  

These complete the modeling of the longitudinal and transverse muscles.  The formulae for muscle stretch strain, forces, and couples are reported in Table~\ref{tab:model}.

\begin{table*}[!tb]
	\caption{Summarized muscle model}
	\centering
	\footnotesize
	\begin{tabular}{c;{3.14159pt/2.71828pt}ccccc;{3.14159pt/2.71828pt}ccc} 
	    \hline
		& \multicolumn{5}{c;{3.14159pt/2.71828pt}}{Muscle model notation} &  \multicolumn{3}{c}{Parameter values for simulation}\\
		\hline
		\hline
		&&&&&&&\\[-0.8em]
		$\muscle$ & $r^\muscle$ & $\nu^\muscle$ & $n^\muscle$ & $m^\muscle$ & $W^\muscle$ & off center & max stress & cross sectional \\
		muscle & position& muscle strain & muscle force & muscle couple & muscle stored energy & distance & [kPa] & area \\
		\hline
		&&&&&&&\\[-0.4em]
		$\LMt$ & $\phi^\longitudinalmuscle\mathsf{b}$ & $\nu_1 - {\phi}^\longitudinalmuscle \kappa$ & $ \abs{{n}^{\LMt}} \mathsf{a}$ & $-  {\phi}^\longitudinalmuscle \abs{{n}^\LMt}$ & $n^\longitudinalmuscle_{\text{max}} A^\longitudinalmuscle(s) F_l (\nu^{\LMt})$ & \multirow{3}{*}{${\phi}^\longitudinalmuscle = \dfrac{2\phi(s)}{3}$} & \multirow{3}{*}{$n^\longitudinalmuscle_\text{max}=19.89$} & \multirow{3}{*}{$A^\longitudinalmuscle =\dfrac{A}{9}$} \\
		&&&&&&&&\\[-0.4em]
		&&&&&&&&\\[-0.6em]
		$\LMb$ & $-\phi^\longitudinalmuscle\mathsf{b}$ & $\nu_1 + {\phi}^\longitudinalmuscle \kappa$ & $\abs{{n}^{\LMb}} \mathsf{a}$ & $ {\phi}^\longitudinalmuscle \abs{{n}^{\LMb}}$ & $n^\longitudinalmuscle_{\text{max}} A^\longitudinalmuscle(s) F_l (\nu^{\LMb})$ & & &  \\
		&&&&&&&&\\[-0.6em]
		\hline
		&&&&&&&&\\[-0.4em]
		$\transversemuscle$ & $0$ & $2 - \tfrac{\nu_1}{\nu_1^\circ}$ & $-\abs{{n}^{\transversemuscle} }\mathsf{a}$ & 0 & $ n^\transversemuscle_{\text{max}} A^\transversemuscle (s) F_l (\nu^\transversemuscle)$ & ${\phi}^\transversemuscle=0$ & $n^\transversemuscle_\text{max}=13.26$ & $A^\transversemuscle = \dfrac{A}{12}$ \\
		&&&&&&&\\[-0.6em]
		\hline
	\end{tabular}
	\label{tab:model}
	\vspace{-10pt}
\end{table*}

\subsection{Muscle energy function} 
\label{sec:muscle energy}

The main result of this section is to obtain the muscle forces and couples $(n^\muscle, m^\muscle)$ from a stored energy function.  The proof of the  following proposition appears in Appendix~\ref{appdx:muscle_energy_proof}.  

\medskip

\begin{proposition} \label{prop:muscle_energy}
Suppose $u^\muscle  = u^\muscle (s, t) $ is a given activation of a generic muscle $\muscle$.  Then there exists a function $W^{\muscle}=W^{\muscle}(\nu_1,\nu_2,\kappa)$ such that the internal muscle forces and torques are given by
\begin{align}
	n^\muscle_i =u^\muscle  \frac{\partial W^\muscle}{\partial \nu_i}  , ~ i = 1, 2,  \quad m^\muscle =u^\muscle \frac{\partial W^\muscle}{\partial \kappa} 
\label{eq:muscle_energy}
\end{align}
The expressions of the function $W^\muscle$ for the two longitudinal muscles $\LMt$ and $\LMb$, and the one transverse muscle $\transversemuscle$ appear in Table \ref{tab:model}.  
\end{proposition}

\medskip
The total stored energy function for the arm is defined as follows:
\begin{align*}
W (\deformations; u) := W^e (\deformations) + \sum\limits_\muscle u^\muscle W^\muscle (\deformations)
\end{align*}
where $u = \{u^\muscle \}$. This results in the potential energy
\begin{align}
\begin{split}
&\potential (u) := \int_0^{L_0} W(w;u) ~\dif s =  \potential^e + \sum\limits_\muscle \potential^{\muscle} (u^\muscle) \\
&\text{where} ~~ \potential^{\muscle} ({u^\muscle}) := \int_0^{L_0} u^\muscle (s,t) W^\muscle (\deformations(s,t))  ~ \dif s
\end{split}
\label{eq:potential_total}
\end{align}
\normalsize

Using this notation, we obtain the following corollary:

\medskip

\begin{corollary} \label{cor:closed_loop_hamiltonian}
Suppose $u^\muscle  = u^\muscle (s, t) $ is a given activation of the muscle $\muscle$.  Then the muscle-actuated arm is a Hamiltonian control system with the Hamiltonian 
\[
\Hamiltonian^{\text{total}} = \kinetic + \potential^e + \sum\limits_{\muscle} \potential^{\muscle} (u^\muscle)
\]
In particular, the muscle forces  in~\eqref{eq:control_system} are given by 
\begin{align*}
\Gmuscle^\muscle (\states) u^\muscle = - \frac{\delta \potential^{\muscle}}{\delta \states}
\end{align*}
\end{corollary}
\medskip

 We omit the proof of Corollary~\ref{cor:closed_loop_hamiltonian} due to lack of space. The proof proceeds in the same way as showing the passive elasticity term in \eqref{eq:control_system} is the (negative) gradient of the arm's intrinsic potential energy function \cite{antman1995nonlinear}. 

How to design the muscle activation $u^\muscle$ to solve a control problem is the main question addressed in the following section on energy shaping.

\section{Energy shaping control design} \label{sec:control}

\subsection{Energy shaping control law} \label{sec:energy_shaping}
In order to obtain and analyze the equilibrium (stationary point of the potential energy), the following definition is useful:
\medskip
\begin{definition} 
Suppose $u^\muscle  = \alpha^\muscle (s) $ is a given time-independent activation of the muscle $\muscle$ and $\alpha=\{\alpha^\muscle\}$.  Then the gradient and the Hessian of $W$ are denoted as
\small
\begin{align} 
P(\deformations; \alpha) := \frac{\partial W (\deformations; \alpha)}{\partial \deformations}, \quad Q(\deformations; \alpha) := \frac{\partial P}{\partial \deformations} = \frac{\partial^2 W(\deformations; \alpha)}{\partial \deformations^2}
\end{align}
\normalsize
\end{definition}

\medskip
The following proposition prescribes an energy shaping control law.

\begin{proposition}\label{prop:control_law}
Consider the control system \eqref{eq:control_system} with a constant muscle control 
\begin{align}
u^\muscle(s,t)  \equiv \alpha^{\muscle} (s),\quad t\geq 0
\label{eq:energy_shaping_control}
\end{align}
Suppose the potential energy functional $\potential(\alpha)$ has a minimum at the deformations $\deformations_\alpha$ with associated configuration $\states_\alpha$. 
Additionally, let the muscle parameters be such that the Hessian $Q(\deformations; \alpha)$ is locally positive definite at $\deformations = \deformations_\alpha$. Then the control law \eqref{eq:energy_shaping_control} renders the equilibrium $\left(\states_\alpha, 0\right)$ of \eqref{eq:control_system} (locally) asymptotically stable.
\end{proposition}

\smallskip
A sketch of the proof of Proposition~\ref{prop:control_law} is given as follows. According to Corollary \ref{cor:closed_loop_hamiltonian}, application of the control law \eqref{eq:energy_shaping_control} makes the closed loop system a damped Hamiltonian system with closed loop Hamiltonian $\Hamiltonian^{\text{total}} = \kinetic + \potential(\alpha)$. The local convexity assumption of $W(w; \alpha)$ lets us take the closed loop Hamiltonian ${\Hamiltonian}^{\text{total}}$ as a Lyapunov functional. Then along a solution trajectory of \eqref{eq:control_system} with controls \eqref{eq:energy_shaping_control}, we have
\begin{align*}
\frac{\dif {\Hamiltonian}^{\text{total}}}{\dif t} = - \gamma \norm{M^{-1} p}^2 \leq 0
\end{align*}
where the norm is taken in the $L^2$ sense. We thus have that the total energy of the system is non-increasing. Finally, an application of the LaSalle's theorem guarantees local asymptotic stability to the largest invariant subset of $\left\lbrace (\states, \momentums) ~\big| ~ \tfrac{\dif {\Hamiltonian}^{\text{total}}}{\dif t} = 0 \right\rbrace$, which is indeed the equilibrium point $(\states_\alpha, 0)$. Note that a complete proof of Proposition~\ref{prop:control_law} involves rigorous arguments of a LaSalle's principle in the infinite dimensional setting, which is beyond the scope of this paper.

\begin{remark}
In general, for underactuated Hamiltonian control systems, the energy shaping methodology requires solving a PDE called the \textit{matching condition} \cite{bloch2000controlled, blankenstein2002matching, ortega2002stabilization}. Our formulation of the muscle model enables us to write the control terms $\Gmuscle^\muscle u^\muscle$ as Hamiltonian vector fields (Corollary~\ref{cor:closed_loop_hamiltonian}). Consequently, the constant control law \eqref{eq:energy_shaping_control} is a solution to the resulting matching condition for this problem. This line of thinking is elucidated in \cite{maschke2000energy}.  
\end{remark}

Thus it remains to design the static controls $\alpha$ that shape the potential energy. This is the subject of the next subsection.

\subsection{Design of potential energy: The static problem} \label{sec:desired_potential}


The problem of designing the muscle activation $\alpha$ is posed as an optimization problem:
\begin{align*}
\begin{split}
\underset{\alpha\in [\text{admissible set}]}{\text{minimize}} \quad &\left[\text{muscle related cost} \right] + \left[\text{task related cost}\right] \\
\text{subject to} \quad &\text{equilibrium constraints (of the rod)} \\
\text{and} 			\quad &\text{task related constraints} 
\end{split}
\end{align*}
This optimization problem is an example of a bilevel optimization problem, also referred to as structural optimization problem in literature \cite{colson2007overview, outrata2013nonsmooth}. 


\medskip
\noindent
\textbf{1) Lower level optimization -- obtain the equilibrium constraints:} This is an example of a forward problem. For a given (fixed) $\alpha$, obtain the equilibrium (or a static) configuration of the rod.  
The equilibrium is obtained by calculating the minimum of the total potential energy $\potential(\alpha)$~\cite{outrata2013nonsmooth} as follows:
\begin{align}
		\begin{split}
			\underset{\deformations(\cdot)}{\text{minimize}}   ~ \potential =& \int_0^{L_0} W(w(s); \alpha) ~ \dif s \\
			\text{ subject to \eqref{eq:state_kinematics},} ~  &\text{ with boundary conditions}
		\end{split}
		\label{eq:statics_given_alpha}
\end{align}
where we recall~\eqref{eq:state_kinematics} is the kinematic constraint of the rod.  

The necessary conditions for optimality are obtained from the Pontryagin's Maximum Principle (PMP) as follows: Denote the costate to $\states(s)$ as $\lambda(s) = (\lambda_1 (s), \lambda_2 (s), \lambda_3(s))^\transpose \in \R^3$. Then the control Hamiltonian
for this problem is
\begin{align*}
H(\states, \lambda, \deformations) = \lambda^\transpose g (\states, \deformations) - W(\deformations; \alpha)
\end{align*}
The costate $\lambda$ evolves according to Hamilton's equation
\begin{align}
\partial_s \costates = - \frac{\partial H}{\partial q} = \begin{bmatrix}
0 \\ 0 \\ \left\lbrace -\nu_1 (-\lambda_1 \sin \theta + \lambda_2 \cos \theta) \right. \\
\left. \nu_2 (\lambda_1 \cos \theta + \lambda_2 \sin \theta ) \right\rbrace
\end{bmatrix} 
\label{eq:costates_evolution}
\end{align}
\normalsize
Pointwise maximization of the control Hamiltonian leads to the requirement that
\begin{align}
\left(\frac{\partial g}{\partial w}\right)^{\transpose} \lambda - P(\deformations; \alpha) = 0
\label{eq:static_strains_general}
\end{align}
In general settings,~\eqref{eq:state_kinematics},~\eqref{eq:costates_evolution} and~\eqref{eq:static_strains_general} together with the appropriate boundary conditions (see e.g.~\cite{bretl2014quasi, chang2020energy}) represent the equilibrium constraint obtained from the lower level optimization problem.  Considerable simplification arises when the boundary conditions are of the fixed-free type.  This case is of particular interest for the \textit{CyberOctopus} control problem where the arm is attached to the head, i.e. the base is fixed (and without loss of generality can be taken as 0). The tip is free since there is no externally imposed boundary condition at the tip. The result is described in the following proposition:\\
\vspace{-5pt}

\begin{proposition} \label{prop:static_strains_hills}
Consider the optimization problem \eqref{eq:statics_given_alpha} with the boundary conditions $\states(0) = 0$ and $\states(L_0)$ free. Then 
any minimizer, denoted by $w_\alpha (s) = (\nu_{1,\alpha}(s), \nu_{2,\alpha}(s), \kappa_\alpha(s))$, must satisfy for all $s \in [0, L_0]$ 
\small
\begin{align}
P(\deformations_\alpha; \alpha) = 
\begin{bmatrix} EA ({\nu}_{1,\alpha} (s)- 1) \\ GA {\nu}_{2,\alpha}(s) \\ EI {\kappa}_\alpha (s)\end{bmatrix} 
+\begin{bmatrix}
\sum  n^{\muscle}_1 (s, {w}_\alpha(s); \alpha^\muscle)  \\ 0 \\ \sum  m^{\muscle} (s, {w}_\alpha(s); \alpha^\muscle) \end{bmatrix} = 0
\label{eq:static_strains_hills}
\end{align} 
\normalsize
\end{proposition}
\vspace*{5pt}
\begin{proof}
Indeed, when $\states(L_0)$ is free, the PMP equations need to be augmented by the transversality condition $\lambda(L_0) = 0$ which, by the virtue of costate evolution equations \eqref{eq:costates_evolution}, leads to $\lambda \equiv 0$ for all $s$. Equation \eqref{eq:static_strains_general} then simplifies to \eqref{eq:static_strains_hills} at $w = w_\alpha$. 
\end{proof}

\smallskip
In summary, equation~\eqref{eq:static_strains_hills} is the equilibrium constraint from solving the forward problem (for a given $\alpha$).  

\newcommand{\insertbox}[1]{%
  \rlap{\smash{\parbox[b]{7cm}{#1}}}%
}
	\begin{algorithm}[t]
		\begin{algorithmic}[1]
			\Require Task (reaching, grasping)
			\Ensure Optimal activations $\bar{\alpha}=(\bar{\alpha}^{\longitudinalmuscle_\text{t}},\bar{\alpha}^{\longitudinalmuscle_\text{b}},\bar{\alpha}^\transversemuscle)$
			\State Initialize: activations $\alpha^{(0)}$, states at base: $\states(0) = 0$ 
			\For{$k=0$ to MaxIter}
			\State Solve \eqref{eq:static_strains_hills} to obtain $\deformations^{(k)}_\alpha$
			\State Update forward \eqref{eq:state_kinematics}
			\State Update backward \eqref{eq:transversality_condition_higher},~\eqref{eq:costates_evolution_higher}
			\State Update activations: $\alpha^{(k+1)} = \alpha^{(k)}+\eta\frac{\partial \hat{\controlHamiltonian}}{\partial \alpha}$
			\State Limit activations $\alpha^{(k)}$ within $[0, 1]$
			\EndFor
			\State Output the final activations as $\bar{\alpha}$ \insertbox{\hspace{-30pt} $\left.\begin{array}{c}~\end{array}\right]$\vspace{70pt}}\insertbox{\hspace{-13pt} \rule{6pt}{0.2pt}  \vspace{75.5pt}} \insertbox{\hspace{-5pt} lower\vspace{77pt}} \insertbox{\hspace{-7pt} level\vspace{67pt}} \insertbox{\hspace{26pt} $\left.\begin{array}{c}~\\~\\~\\~\\~\\~\\~\end{array}\right]$\vspace{8pt}}\insertbox{\hspace{44pt} \rule{6pt}{0.2pt}  \vspace{52.5pt}} \insertbox{\hspace{50pt} higher\vspace{54pt}} \insertbox{\hspace{50pt} level\vspace{44pt}}
		\end{algorithmic}
		\caption{Solving the bilevel optimization problem}
		\label{alg:forward_backward}
	\end{algorithm}
\setlength{\textfloatsep}{5pt}

\vspace{10pt}
\noindent
\textbf{2) Upper level optimization -- obtain the optimal static actuation:} This is an example of an inverse problem. For a given set of  control-specific tasks and constraints, obtain the activation $\alpha$ that solves the task.  For this purpose, we propose the following optimization problem: 
\small
\begin{align}\label{eq:optimization_alpha}
	\begin{split}
	     \underset{\alpha(\cdot),\; \alpha^\muscle (s) \in [0,1]}{\text{minimize}} &\quad \mathsf{J}(\alpha) =\frac{1}{2}\int_0^{L_0} \sum\limits_\muscle \left(\alpha^\muscle (s) \right)^2  
	    ~\dif s \\
		& \hspace*{-40pt}+  \int_0^{L_0} \mu_{\text{grasp}}(s) \Phi_{\text{grasp}} (\states (s)) ~ \dif s  +\mu_{\text{tip}}\Phi_{\text{tip}}({q}(L_0),q^*)\\
		\text{subject to} &\quad  \partial_s {q}=g(q, {w_\alpha}), ~~ {q}(0)=0, ~ q (L_0) ~ \text{free}; \\
		\text{and} &\quad \Psi_j (\states) \leq 0, ~~ j = 1, 2, ..., N_{\text{obs}}
	\end{split}
\end{align}
\normalsize
The significance of the terms is as follows:
\begin{enumerate}
\item The quadratic term is used to model the control cost of using muscles.
\item The function $\Phi_{\text{tip}}$ is used to model the control task for the tip, e.g., to reach a given point in space.
\item The running cost $\Phi_{\text{grasp}}$ is used to model the control task for the whole arm, e.g., for the arm to wrap around an object.  
\item Equation~\eqref{eq:optimization_alpha} is used to model the obstacles in the environment as state constraints of the form $\Psi(\states) \leq 0$.
\end{enumerate}
The formulae for these functions and parameters (e.g. $\mu_{\text{grasp}}, \mu_{\text{tip}}$) are task-specific and appear in Sec.~\ref{sec:numerics}. The higher level optimization problem \eqref{eq:optimization_alpha} is an optimal control problem for the kinematics  \eqref{eq:state_kinematics} with $\alpha$ as controls, whose necessary conditions for optimality are obtained by PMP and are summarized in Appendix~\ref{appdx:pmp_higher}. 


\medskip
\noindent
\textbf{3) Algorithms:}
Given any task, we solve the bilevel optimization problem in an iterative manner. In each iteration, we first solve the lower level problem, i.e. solve the nonlinear equations \eqref{eq:static_strains_hills} for $\deformations_\alpha$ pointwise in $s$. We utilize the \textit{fsolve} routine in the scipy package for this purpose. The higher level problem \eqref{eq:optimization_alpha} is then solved by using a forward-backward algorithm to obtain the optimal $\alpha$ (see also Sec. III-C in \cite{chang2020energy}). In Algorithm~\ref{alg:forward_backward}, we give a brief pseudo code of the algorithm. Finally according to Proposition~\ref{prop:control_law} the energy shaping controls are simply the optimal $\alpha$. 

A more comprehensive discussion on algorithms for solving this problem appears in our prior work~\cite{chang2020energy}.   


\begin{figure*}[!ht]
	\centering
	\includegraphics[width=\textwidth , trim = {30 480 30 70}, clip = false]{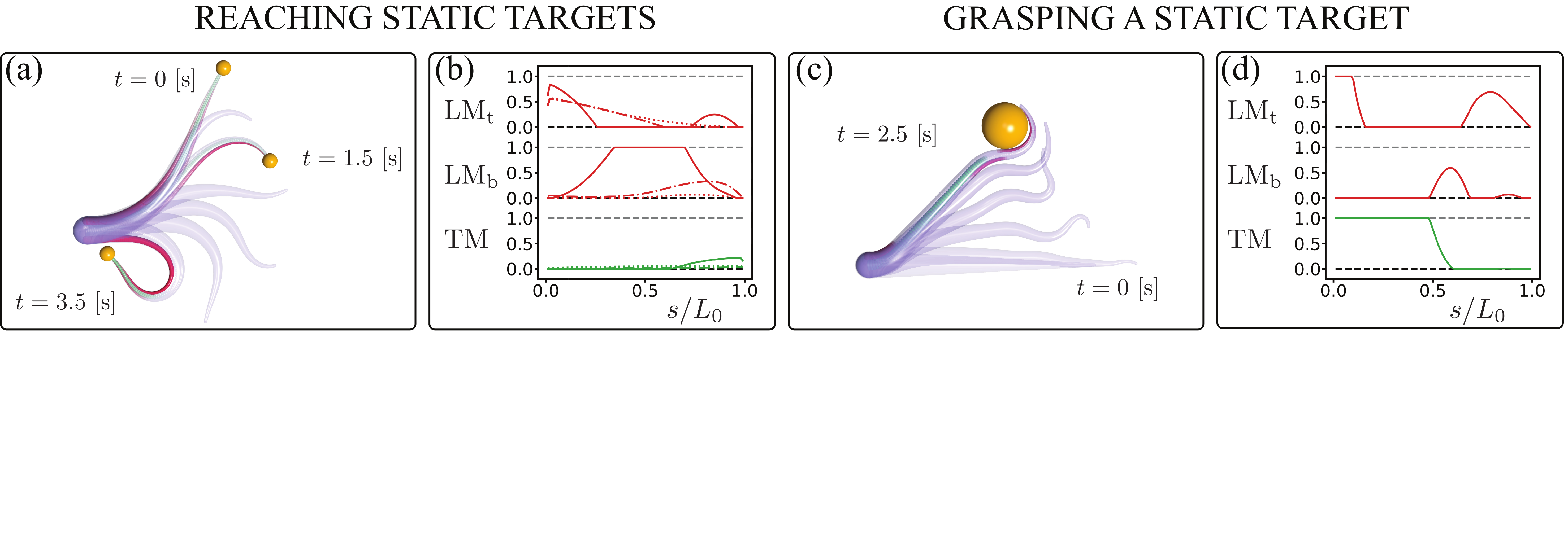}
	\caption{Reaching task (a-b). The arm is tasked to reach three different locations one after the other, mimicking
the octopus' fetching motions. (a) Targets are located at $(12, 14)$, $(16,6)$ and $(2,-2)$ [cm] and are indicated as orange spheres. Optimal arm configurations are depicted together with muscle activations. The time evolution of the arm is depicted as transparent purple rods. (b) Dotted lines show the activations at $t=0$ [s]; dash-dot lines show the activations at $t=1.5$ [s]; and solid lines show the activations at $t=3.5$ [s]. Grasping task (c-d). The arm is tasked to wrap around a static sphere (the big orange sphere) of radius $2$ [cm] centered at $(12, 12)$ [cm]. (d) The solid line shows the activations at $t=2.5$ [s].}
	\label{fig:experiments}
	\vspace{-10pt}
\end{figure*}

\section{Numerical simulation}\label{sec:numerics}
In this section, a numerical environment is used to demonstrate the abilities of a soft octopus arm under our control algorithm. Two experiments are shown to mimic the behaviors of reaching and grasping motion of an octopus arm. 
\subsection{Simulation setup}
The explicit dynamic equations of motion \eqref{eq:control_system} of a planar Cosserat rod \cite{antman1995nonlinear} are discretized into $N_\text{d}$ connected cylindrical segments and solved numerically by using our open-source, dynamic, three-dimensional simulation framework \textit{Elastica} \cite{gazzola2018forward, zhang2019modeling, naughton2021elastica}. 
The radius profile of the tapered geometry of an octopus arm is modeled as \[\phi(s)=\phi_{\text{tip}}s+\phi_{\text{base}}(L_0-s)\] with the cross sectional area and the second moment of area as $A=\pi\phi^2$ and $I=\frac{A^2}{4\pi}$. 
We take the rest configuration of the rod to be straight of length $L_0$, i.e. $(\nu_1^\circ, \nu_2^\circ, \kappa^\circ) \equiv (1, 0, 0)$.
The biologically realistic parameter values in this work are listed in Table \ref{tab:model} and \ref{tab:parameters} and also in reference \cite{chang2020energy}. The following force length curve $f_l(\cdot)$ for the Hill's muscle model is used:
\begin{equation*}
	\footnotesize
	f_l(z)=\left\{\begin{array}{ll}
		3.06 z^3 - 13.64 z^2 + 18.01 z - 6.44, & \text{if }0.6\leq z\leq1.6\\
		0, &\text{else}
		\end{array}\right.
\end{equation*}
This model is fitted from experimental data~\cite[Fig. 6]{kier2002fast}.

\subsection{Experiments}

\subsubsection{Reaching multiple static targets}
The first experiment consists reaching multiple targets one by one to mimic the behavior of a real octopus. For any target, the tip cost function $\Phi_{\text{tip}}$ in \eqref{eq:optimization_alpha} is set as
\begin{equation*}
\Phi_{\text{tip}}(q(L_0),r^*)=\frac{1}{2}\abs{r^*-r(L_0)}^2
\end{equation*}
where $r^*$ is the target position. The cost does not depend on the tip angle $\theta(L_0)$ as we are not concerned about the tip pose. The weight function $\mu_{\text{grasp}}$ is chosen to be zero and regularization parameter $\mu_{\text{tip}}$ is set as $10^5$. Simulation results are shown in Fig.~\ref{fig:experiments}(a-b). 

\subsubsection{Grasping an object}
In the second experiment, the octopus arm is tasked to grasp a target sphere. This behavior is commonly seen when an octopus is trying to reach for a bottle, a shell or a crab. To find the desired static configuration via \eqref{eq:optimization_alpha}, the object is treated as both an obstacle and a target so that the arm cannot penetrate it but can wrap around it. The inequality constraint model of it is:
\begin{equation*}
\Psi(\states(s))=\left(\phi^\text{obj}+\phi(s)\right)^2-|r^\text{obj}-r(s)|^2
\end{equation*}
where $\phi^\text{obj}$ and $r^\text{obj}$ denote the radius and center position of the object, respectively. In addition, since we want the arm to grasp the target sphere, the running cost $\Phi_{\text{grasp}}$ and the weight function $\mu_\text{grasp}$ in \eqref{eq:optimization_alpha} are designed so that the arm can get as close to the boundary as possible:
\begin{equation*}
\Phi_{\text{grasp}}(\states(s)) = \text{dist}(\Omega, r(s)), ~~ \mu_{\text{grasp}}(s)= 10^5 \chi_{_{[0.4L_0,L_0]}} (s) 
\end{equation*}
where $\Omega$ denotes the boundary of the object (here just a circle), $\text{dist}(\cdot, \cdot)$ calculates the distance between the boundary and the point $r(s)$, and  $\chi_{_{[s_1,s_2]}} (\cdot)$ denotes the characteristic function of $[s_1,s_2]$. 
Such a design together with the inequality constraint cause the distal portion of the arm, starting from $s = 0.4L_0$, to grasp the target sphere without penetrating it. The value of $\mu_{\text{tip}}$ is 0 in this case. Fig. \ref{fig:experiments}(c-d) shows the simulation results where the arm grasps the sphere under its muscle actuation model.

\begin{table}[tb]
	\centering
	\caption{Parameters}
	\begin{tabular}{clc}
		\hline
		\hline\noalign{\smallskip}
		Parameter & Description & Value \\
		\hline\noalign{\smallskip}
		$L_0$ & rest arm length [cm]& $20$ \\
		$\phi_\text{base}$ & base radius [cm] & $1.2$\\
		$\phi_\text{tip}$ & tip radius [cm] & $0.12$\\
		$E$ & Young's modulus [kPa] & $10$ \\
		$G$ & Shear modulus [kPa] & $40/9$ \\
		$\rho$ & density [kg/m$^3$] & $1050$ \\
		$\gamma$ & dissipation [kg/s] & $0.02$ \\
		$N_\text{d}$ & discrete number of elements & $100$ \\
		$\Delta t$ & discrete time-step [s] & $10^{-5}$ \\
		$\eta$ & learning rate & $10^{-8}$ \\
		\hline
	\end{tabular}
	\label{tab:parameters}
\end{table}

\section{Conclusion and future work} \label{sec:conclusion}
In this paper, a flexible octopus arm is represented as a planar Cosserat rod and its muscle mechanisms are modeled as distributed internal force/couple functions with the muscle activations as the control inputs. The rod is viewed as an underactuated Hamiltonian control system, for which an energy shaping control method is sought to solve various manipulation objectives, e.g. reaching and grasping. We have shown that the total energy of the closed loop system can be expressed by augmenting the inherent elastic energy of the rod with muscle stored energy functions. As a result, constant muscle controls stabilize the arm. A bilevel optimization problem is then constructed and solved numerically to obtain desired muscle controls for a given task. Numerical experiments demonstrate the efficacy of this scheme. As a direct extension, a more sophisticated muscle actuation model and the corresponding control method can be applied to the general 3D case. Another direction of future work is to develop an octopus-inspired neuromuscular control where muscle activation is controlled by underlying neuronal activity. 
The problem of sensorimotor control can also be considered where only a part of the state is available to the controller through internal distributed sensors.

\bibliographystyle{IEEEtran}
\bibliography{reference}

\appendices
\renewcommand{\thelemma}{A-\arabic{section}.\arabic{lemma}}
\renewcommand{\thetheorem}{A-\arabic{section}.\arabic{theorem}}
\renewcommand{\theequation}{A-\arabic{equation}}
\renewcommand{\thedefinition}{A-\arabic{definition}}
\setcounter{lemma}{0}
\setcounter{theorem}{0}
\setcounter{equation}{0}


\section{Proof of Proposition \ref{prop:muscle_energy}} \label{appdx:muscle_energy_proof}
\begin{proof}
Indeed, for the top longitudinal muscle $\LMt$, define $W^{\LMt} := A^{\longitudinalmuscle} n^{\longitudinalmuscle}_{\text{max}} F_l (\nu_1 - \phi^{\longitudinalmuscle} \kappa)$.
Then it readily follows by using the definition of the function $F_l$ that
\begin{align*}
n^{\LMt}_1 &= \abs{n^{\LMt}} = u^{\LMt} \frac{\partial W^{\LMt}}{\partial \nu_1} \\
m^{\LMt}    &= - \phi^{\longitudinalmuscle} \abs{n^{\LMt}} = u^{\LMt} \frac{\partial W^{\LMt}}{\partial \kappa} 
\end{align*}
Additionally, it is obvious that $n^{\LMt}_2 = \frac{\partial W^{\LMt}}{\partial \nu_2} = 0$. Similar arguments follow for other two muscles. 
\end{proof}

\section{PMP conditions for problem \eqref{eq:optimization_alpha} } \label{appdx:pmp_higher}
Here we only express the key PMP conditions. For more explanations, please refer to \cite{chang2020energy}. For the constrained optimization problem \eqref{eq:optimization_alpha}, the original state $\states$ is augmented to include $N_{\text{obj}}$ additional states $\hat{q}_j$ which has the following evolution:
\begin{equation*}
	\partial_s\hat{q}_j=c_j(q)=\max(\Psi_j(q),0),~~ \hat{q}_j(0)=0, ~~ j = 1,..., N_{\text{obs}}
\end{equation*}
and the terminal cost function is also modified as follows
\begin{equation*}
	\hat{\Phi}(q(L_0)) = \mu_\text{tip}\Phi_\text{tip}(q(L_0), q^*)+\sum\xi_j\hat{q}_j(L_0)
\end{equation*}
where $\xi_j>0$ are the weights for the augmented states $\hat{q}_j(L_0)$. Denoting the costate of $q$ as $\hat{\lambda}$, the modified control Hamiltonian $\hat{H}$ is written as
\begin{equation*}
	\hat{H}(s, q, \hat{\lambda},\alpha) = \hat{\lambda}^\transpose g (\states, \deformations_\alpha) - \frac{1}{2} \abs{\alpha}^2-\mu_\text{grasp}(s)\Phi_\text{grasp}(q)
\end{equation*}
The costate $\hat{\lambda}$ must satisfy the Hamilton's equation
\begin{align}\label{eq:costates_evolution_higher}
	\partial_s\hat{\lambda}=-\frac{\partial\hat{H}}{\partial q}+\sum\xi_j\frac{\partial c_j}{\partial q}
\end{align}
with the accompanied transversality condition
\begin{equation}\label{eq:transversality_condition_higher}
	\hat{\lambda} (L_0) = - \mu_{\text{tip}} \frac{\partial \Phi_{\text{tip}}}{\partial \states} (q(L_0), q^*)
\end{equation}
and the optimal $\alpha$ should maximize the Hamiltonian $\hat{H}$ pointwise in $[0, L_0]$.

\end{document}